\documentclass[letterpaper]{article} 
\usepackage{aaai25}  
\usepackage{times}  
\usepackage{helvet}  
\usepackage{courier}  
\usepackage[hyphens]{url}  
\usepackage{graphicx} 
\usepackage{natbib}  
\usepackage{caption} 
\usepackage{algorithm}
\usepackage{algorithmic}
\usepackage{booktabs}
\usepackage{amssymb}
\usepackage{amsmath}
\usepackage{multirow}
\usepackage{array}

\usepackage{newfloat}
\usepackage{listings}
\DeclareCaptionStyle{ruled}{labelfont=normalfont,labelsep=colon,strut=off} 
\floatstyle{ruled}
\newfloat{listing}{tb}{lst}{}
\floatname{listing}{Listing}
\title{Multi-View Incremental Learning with Structured Hebbian Plasticity for Enhanced Fusion Efficiency}
\author{
    Yuhong Chen\textsuperscript{\rm 1,}\textsuperscript{\rm 2},
    Ailin Song\textsuperscript{\rm 2,}\textsuperscript{\rm 3},
    Huifeng Yin\textsuperscript{\rm 4},
    Shuai Zhong\textsuperscript{\rm 2},
    Fuhai Chen\textsuperscript{\rm 1},
    Qi Xu\textsuperscript{\rm 5},\\
    Shiping Wang\textsuperscript{\rm 1}\thanks{Corresponding Authors.},
    Mingkun Xu\textsuperscript{\rm 2,}\textsuperscript{\rm 4}\footnotemark[1]
}
\affiliations{
    \textsuperscript{\rm 1}College of Computer and Data Science, Fuzhou University, Fuzhou, China\\
    \textsuperscript{\rm 2}Guangdong Institute of Intelligence Science and Technology, Hengqin, Zhuhai, China\\
    \textsuperscript{\rm 3}University of Electronic Science and Technology of China, Chengdu, China\\
    \textsuperscript{\rm 4}Center for Brain Inspired Computing Research, Department of Precision Instrument, Tsinghua University, Beijing, China\\
    \textsuperscript{\rm 5}School of Computer Science and Technology, Dalian University of Technology, Dalian, China

    yhchen2320@163.com, sal314159@163.com, yinhf22@mails.tsinghua.edu.cn, zhongshuai@gdiist.cn, chenfuhai3c@163.com, xuqi@dlut.edu.cn, shipingwangphd@163.com, xumingkun@gdiist.cn
}

\usepackage{bibentry}

\begin{document}

\maketitle

\begin{abstract}
The rapid evolution of multimedia technology has revolutionized human perception, paving the way for multi-view learning. However, traditional multi-view learning approaches are tailored for scenarios with fixed data views, falling short of emulating the intricate cognitive procedures of the human brain processing signals sequentially. Our cerebral architecture seamlessly integrates sequential data through intricate feed-forward and feedback mechanisms. In stark contrast, traditional methods struggle to generalize effectively when confronted with data spanning diverse domains, highlighting the need for innovative strategies that can mimic the brain's adaptability and dynamic integration capabilities. In this paper, we propose a bio-neurologically inspired multi-view incremental framework named MVIL aimed at emulating the brain's fine-grained fusion of sequentially arriving views. MVIL lies two fundamental modules: structured Hebbian plasticity and synaptic partition learning. The structured Hebbian plasticity reshapes the structure of weights to express the high correlation between view representations, facilitating a fine-grained fusion of view representations. Moreover, synaptic partition learning is efficient in alleviating drastic changes in weights and also retaining old knowledge by inhibiting partial synapses. These modules bionically play a central role in reinforcing crucial associations between newly acquired information and existing knowledge repositories, thereby enhancing the network's capacity for generalization. Experimental results on six benchmark datasets show MVIL's effectiveness over state-of-the-art methods.
\end{abstract}

%

\section{Introduction}

Swift advancements in multimedia technology have fundamentally altered human perception of the world and laid the foundation for advanced analytical methods like multi-view learning~\cite{wen2023efficient, li2021consensus, wang2019parameter, wu2023interpretable, zhuang2024enhancing, lu2024towards}. 
This technology enhances information abundance by describing data in varied formats and views. 
In practical applications, multi-view data are collected and constructed dynamically, especially in cutting-edge fields like brain-computer interfaces~\cite{zhu2023dynamical, kazi2022dg}, which use multiple sensors, such as electrodes, to monitor neural signals and convert them into commands, quite similar to the complex way in which the human brain handles multiple data.  
The human brain perceives and makes decisions by gathering information from diverse sensory channels, such as visual, olfactory, and auditory ones. 
Nevertheless, these sensory signals to the brain's processing center are not instantaneous~\cite{pesnot2022multisensory}.
To compensate for this temporal disparity and seamlessly integrate the diverse streams of information, the brain deploys sophisticated mechanisms that intertwine feed-forward and feedback mechanisms~\cite{hou2019neural, zhang2022unveiling,jiang2023adaptive,wu2022brain}. 
The brain continuously interacts with the external environment sequentially, constantly learning and optimizing its overall perception, thereby achieving a higher level of adaptability and intelligence~\cite{chiel1997brain, ito2022constructing}.
\begin{figure}[!t]
	\centering
	\includegraphics[width=\linewidth]{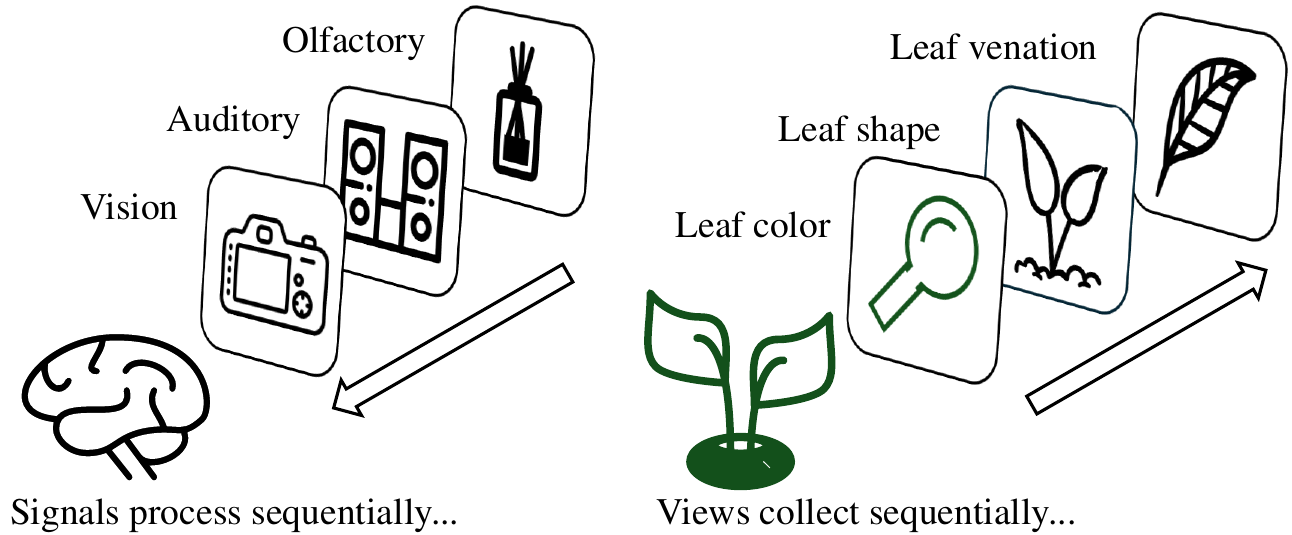}\\
	\caption{Sensory signals reach the brain sequentially, and in parallel with this, the view data increases dynamically.}
	\label{forgetting}
\end{figure}

However, traditional multi-view learning deals with scenarios where all data views are available in advance~\cite{wandecouple, 10462517, wan2023auto}. 
Specifically, they prioritize learning new view data over the whole knowledge accumulation, thus lacking the adaptability and scalability to efficiently integrate sequentially acquired data from multiple views. 
In fact, they neglect the scenarios in which the human brain dynamically handles and conveys new view information, adapts, and optimizes overall cognition.
This contrasts sharply with the synaptic plasticity observed in the human brain.
In neuroscience, synaptic strengthens or weakens itself based on excitation or inhibition between the pre- and post-neurons, which is known as Hebbian theory~\cite{golkar2020simple, triche2022exploration, duan2023hebbian}. 
As information streams in constantly, the feed-forward mechanism deftly extracts pertinent and consistent details, while leveraging the feedback mechanism to refine and rectify this information. This integrated approach fosters a profound interconnection between old and new knowledge, facilitating the comprehensive understanding and enhancing cognitive capabilities~\cite{xiao2023hebbian,shen2024efficient,zhang2019fast}.

As mentioned above, in reality, views are generally added one by one, yet a view carries limited information, which makes it intractable to integrate view associations at a fine-grained scale to make the network stable when the domain changes.
Several works have been done to tackle the dilemma in multi-view continual clustering~\cite{wan2022continual,yan2024live, wan2024fast}.
Some methods learn a consensus weight matrix to preserve the old views' topology~\cite{mitash2017self, cai2023lifelong}. 
However, the views' inherent differences may cause the generated topology to be less comprehensive.
And models are more inclined to have faith in past views which renders it harder to knowledge accumulation.
Moreover, they focus on clustering and cannot tackle classification scenarios.

To tackle the relentless expansion and increasing complexity of data, multi-view continual learning has emerged~\cite{sun2020continual, wang2021continuous}. However, it primarily focuses on scenarios where new classes of data emerge, with each task encompassing 2-3 non-overlapping classes. 
It emphasizes the continual learning effects in different tasks, with the aim that decision-making for old tasks is not interfered with by new knowledge.
However, it overlooks the scenarios of the growth of views to describe data. 
Thus we introduce multi-view incremental learning, which aims to gradually merge the view information to achieve a holistic knowledge of the data by utilizing the consistency and complementarity of multi-view data.
Each task in this framework contains all classes of data, ensuring that no new classes emerge. Critically, each task presents the data from a unique view, so an ongoing fusion of complementary information is urgently needed.
Specifically, we design a biologically plausible module named structured Hebbian plasticity, which facilitates the seamless integration of new incoming view data with existing view representation by reshaping the structure of weights. 
Besides, we devise a synaptic partition learning that depresses some of the synapses to minimize drastic changes in weights.
The former reinforces synapses, while the latter suppresses them. 
The interaction between both mechanisms enables the model to effectively regulate weight changes, thereby mitigating the oblivion of knowledge.
Our method offers an efficient solution for real-world applications.
We conducted extensive experimental evaluations to demonstrate the superior performance of our method for node classification.
The main contributions are as below:
\begin{itemize}
	\item 
 We discern the constraints inherent in traditional multi-view learning when confronted with the reality of increasing view data, i.e., the lack of ability to generalize to new data.
	\item We firstly propose a bio-inspired multi-view incremental learning framework, which designs a structured Hebbian plasticity to reinforce views' associations for efficient knowledge integration of sequentially arriving data.
	\item The proposed method is employed in multi-view semi-supervised classification tasks, showcasing superior performance compared with other state-of-the-art graph-based algorithms.
\end{itemize}

\section{Related Work}
In this section, we present a concise introduction of the most pertinent work pertaining to our study in this paper, including multi-view learning and multi-view continual learning.
\subsection{Multi-view Learning}

Multi-view learning~\cite{wang2015deep, song2023gaf, 10616103} involves integrating and encoding information from multiple sources to derive a low-dimensional representation that captures both consistency and complementarity across views.
It is worth noting that techniques from many emerging fields like graph learning~\cite{chen2024heterogeneous,chen2023agnn,wu2024graph, NEURIPS2023_3ec6c6fc}, have prompted the fast development of multi-view learning methods.
By leveraging the ability to construct a comprehensive understanding of objects, multi-view learning provides versatile solutions for complex data analysis tasks in different domains, such as bioinformatics~\cite{huang2023multi, luo2018multi, thammasorn2021nearest, ijcai2024p277}, social network analysis~\cite{lan2017learning, chen2023dual, ma2017multi,chen2024attributed}, recommender systems~\cite{zou2022multi, wang2020m2grl, cui2018mv}. 
However, conventional multi-view learning methods presuppose that all views are constantly available.
This poses a significant challenge: as data is continuously updated or augmented, the multi-view learning model necessitates retraining, significantly impeding efficiency. 
Therefore, to effectively accommodate the continually expanding number of view data, multi-view learning must possess the ability for continual knowledge retention and generalization.

\begin{figure*}[!ht]
	\centering
	\includegraphics[width=0.8\textwidth]{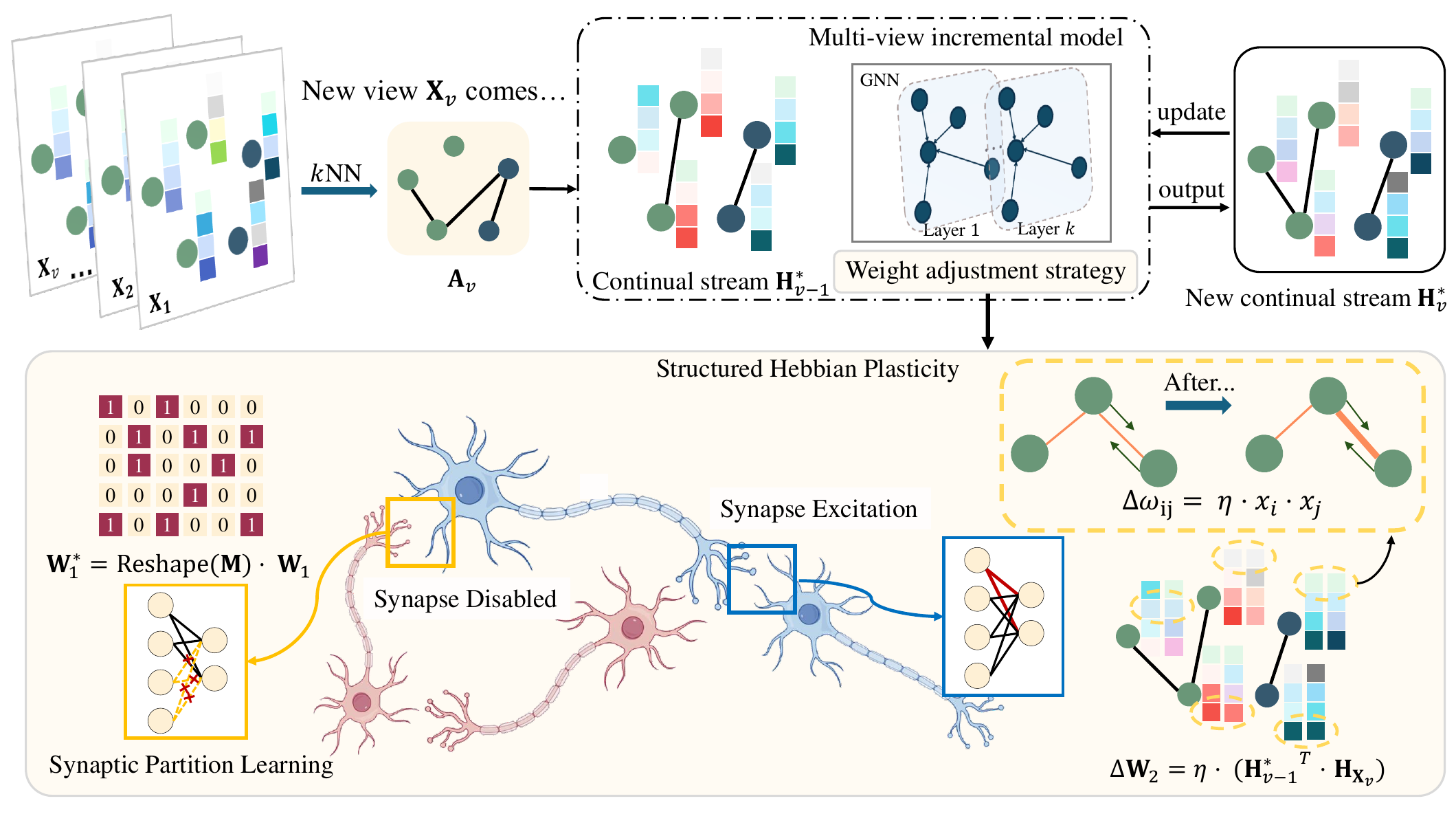}\\
	\caption{An overview of the proposed framework MVIL, which is drawn by Figdraw. MVIL introduces a weight adjustment strategy that incorporates structured Hebbian plasticity and synaptic partition learning, aiming to allow the model to learn with higher biological fidelity. The structured Hebbian plasticity reshapes the weight structure by reinforcing the consistent information between the old and new views; while the latter slows down weight changes by suppressing some of the synapses. 
 }
	\label{Framework1}
\end{figure*}

\subsection{Multi-view Continual Learning}
Current multi-view continual learning has two dominant classifications: task-incremental and class-incremental types~\cite{van2022three}. 
Multi-view task-incremental learning~\cite{li2017lifelong, sun2018robust} establishes connections between tasks and extracts consistent information across multiple views to complement each task learning, thus reducing the forgetting of old tasks as new ones arrive.
Notably, task-incremental learning requires receiving prompts for new tasks, where one task contains multiple views; whereas our view incremental learning does not receive prompts for new tasks, and only one view is included in a task.
However, in practice, the prompts for new tasks are hard to distinguish.
Consequently, detecting the task boundary becomes a formidable task.
Thus, the multi-view class-incremental learning~\cite{yang2022multi, li2024multi} has been proposed, where new class data is seamlessly introduced without explicit prompting to the model, requiring the model to autonomously discern the onset of a new task. These methods focus on regularizing the update of weights, thus alleviating the forgetting of old tasks.
Class-incremental learning assumes that data growth leads to new unknown classes, where a task contains 2-3 classes of data with multiple views, while our view incremental learning assumes that data growth leads to new descriptions, and a task includes all the classes from one view, with no class growth.
We primarily emphasize overall cognitive capture, whereas continual learning exclusively centers on discrete task-specific decisions.

\section{Preilminary}
\begin{table}[!ht]
	\resizebox{\linewidth}{!}{
		\centering
		\begin{tabular}{l||c}
			\toprule {Notations} & {Descriptions} \\ \midrule
			{$ \mathbf{X}_{v} \in \mathbb{R}^{n\times d_v}$}   &  The $v$-th given multi-view feature matrix.\\
			{$ \mathbf{A}_{v} \in \mathbb{R}^{n\times n}$}   &  The similarity matrix evaluated from $ \mathbf{X}^{\left(v\right)} $ by $k$NN.\\
            {$ \mathbf{H}_{v} \in \mathbb{R}^{n\times c}$}   &  The $v$th view initial representation.\\
            {$ \mathbf{H}^{*}_{v} \in \mathbb{R}^{n\times c}$}   & The $v$th view representation after Hebb reinforcement. \\
            {$ \mathbf{M} \in \mathbb{R}^{d_v\times d}$}   & The binary matrix of synaptic partition learning. \\
            {$ \mathbf{W}_{1} \in \mathbb{R}^{d_v\times d}$}   & The first layer weight. \\
            
            {$ \mathbf{W}^{*}_{1} \in \mathbb{R}^{d_v\times d}$}   & The first layer weight after the partition mask. \\
            {$ \mathbf{W}_{2} \in \mathbb{R}^{d\times c}$}   & The second layer weight. \\
            {$ \mathbf{W}^{*}_{2} \in \mathbb{R}^{d\times c}$}   & The second layer weight after Hebb reinforcement. \\
			\bottomrule
	\end{tabular}}
	\caption{The symbolic notations and their descriptions.}
	\label{SymbolicNormalization}
\end{table}
In this section, we start with the introduction of some basic notations used.
Denote given multi-view feature matrices as $\mathcal{X} = {\{\mathbf{X}_v\}_{v=1}^{V}}$, where $\mathbf{X}_v = [\mathbf{x}_1, \cdots , \mathbf{x}_n] \in \mathbb{R}^{n \times d_V}$ is the data from the $v$th view for any $v \in {1,\cdots,V}$. 
$n$ represents the quantity of samples, $V$ signifies the count of views. And $d_v$ indicates the distinct dimensions of the feature matrices, $c$ denotes the number of classes.
Furthermore, to better understand the mathematical notation usages which are primarily used, we list the notation explanations in Table~\ref{SymbolicNormalization}.

\section{Methodology}

As previously mentioned, the major existing methods overlook the scenarios where data observations of new views accumulate gradually over time.  
Consequently, we have devised a multi-view incremental framework MVIL specifically tailored for scenarios where continuously expanding new view data. 
This framework mimics the human brain feed-forward and feedback mechanisms to empower the model to adapt to the ever-growing amount of view data, thereby achieving knowledge accumulation. 
Fig.~\ref{Framework1} provides a comprehensive depiction of our proposed method. 
Notably, we have incorporated a biologically inspired weight adjustment strategy that integrates structured Hebbian plasticity and synaptic partition learning.

\subsection{Streaming View Representation Learning}
Firstly, we design a streaming graph learning model wherein upon the arrival of new view data $\mathbf{X}_{v}$, we construct the corresponding adjacency $\mathbf{A}_{v}$ by applying $k$-Nearest Neighbor ($k$NN). 
When the first view data arrives, we utilize a shared weights two-layer GCN, which can be represented as,
\begin{equation}
    \mathbf{H}_1^{*} = \mathbf{\hat{A}}_{1} \sigma \big( \mathbf{\hat{A}}_{1} \mathbf{X}_{1} \mathbf{W}^{*}_1 \big) \mathbf{W}_2.
\end{equation}
$\mathbf{\hat{A}}_{1} = \mathbf{\widetilde{D}}^{-\frac{1}{2}}_{1}\widetilde{\mathbf{A}}_{1}\mathbf{\widetilde{D}}^{-\frac{1}{2}}_{1}$ and $\widetilde{\mathbf{A}}_{1} = \mathbf{A}_{1} + \mathbf{I}$. 
$\mathbf{A}_{1} \in \mathbb{R}^{n\times n}$ corresponds to an adjacency matrix constructed by $\mathbf{X}_{1}$. 
Herein, $\mathbf{W}^{*}_1 \in \mathbb{R}^{d_v \times d}$ and $\mathbf{W}_2 \in \mathbb{R}^{d \times c}$ denote the learnable shared weight matrices, which would be shared by all the view.
At this stage, the initialization of the streaming model has been completed.
Streaming models continually retain the currently deposited representations without storing the entered data, thus effectively enhancing the privacy of the model.

When the new view data $\mathbf{X}_{v}$ comes, we introduce $\mathbf{X}_{v}$ into the continual stream model and initially integrate the previous knowledge.
As the view $v$ increases, the representation $\mathbf{H}_{v}$ can be expressed as, 
\begin{equation}\label{Hv_initial}
    \mathbf{H}_{v} = \underbrace{ \mathbf{\hat{A}}_{v} \sigma \big( \mathbf{\hat{A}}_{v} \mathbf{X}_{v} \mathbf{W}^{*}_1 \big) \mathbf{W}_2}_{new \; data \; stream } + \underbrace{\alpha \mathbf{H}^{*}_{v-1}}_{incremental \; stream},
\end{equation}
where $v \in \{2,\cdots,V\}$. $\mathbf{\hat{A}}_{v} = \mathbf{\widetilde{D}}^{-\frac{1}{2}}_{v}\widetilde{\mathbf{A}}_{v}\mathbf{\widetilde{D}}^{-\frac{1}{2}}_{v}$ and $\widetilde{\mathbf{A}}_{v} = \mathbf{A}_{v} + \mathbf{I}$. 
$\mathbf{A}_{v} \in \mathbb{R}^{n\times n}$ corresponds to the $v$-th adjacency matrix constructed by particular $\mathbf{X}_{v}$.  $\mathbf{H}^{*}_{v-1} \in \mathbb{R}^{n \times c}$ denotes the representation of the past views when the view $v$ has not been increased. 
$\alpha$ refers to a learnable parameter to balance the weight of the new view data with the past knowledge.
As new data continues to arrive, the streaming model continuously assimilates new knowledge, integrating representation learning.
However, simplistic fusion fails to capture the intricate interactions between views, and the issue of oblivion of knowledge resulting from weight adjustments due to the addition of views remains unaddressed.

\subsection{Structured Hebbian Plasticity}

Inspired by biological neurology, we firstly introduce the Hebbian learning~\cite{golkar2020simple, triche2022exploration} into multi-view graph representation learning. The core of Hebbian's theory is that continuous repetitive stimulation of pre-synaptic and post-synaptic neurons results in an increase in synaptic transmission efficacy. Thus, we design the structured Hebbian plasticity, which facilitates the selective strengthening or weakening of connections based on structured patterns or relations in the data, enhancing the domain's generalization ability.

Specifically, the Hebbian learning rule updates the connection weights between two neurons based on the correlation between them. 
If neuron $i$ and neuron $j$ are active at the same moment, then the connection weight between them will increase. 
Therefore, the rule can be expressed as:
\begin{equation}
    \Delta {w}_{ij} = \eta \cdot x_i \cdot x_j,
\end{equation}
where $\Delta {w}_{ij}$ indicates the amount of change in weights, $\eta$ is the learning rate, $x_i$ and $x_j$ are the outputs of neuron $i$ and neuron $j$.

Instead of uniformly treating all weights, structured Hebbian plasticity selectively identifies biologically plausible node representations based on the similarity between representations across different views, thereby optimizing specific weights, enhancing learning efficiency, and improving generalization performance.
Simultaneous activation of the front and back neurons strengthens the weights connecting this pair of neurons, which coincides with a plausible fusion of the past and new representations.
This happens to be available for enhanced fine-grained representation; if the past and new representations stimulate similar node representations at the same time, we enhance this weight to emphasize the consistent information.
Therefore, we apply it to the weight $\mathbf{W}_2$.

\begin{equation}\label{deltaW}
    \Delta \mathbf{W}_2 = \eta \cdot \mathbf{H}_{new}^{T} \mathbf{H}_{past}, 
\end{equation}
where $\mathbf{H}_{new}^{T} \in \mathbb{R}^{n \times d}$ denotes the new view data and $\mathbf{H}_{past} \in \mathbb{R}^{n \times c}$ refers to the past representation.
Here, we develop Eq.~\eqref{deltaW} in more detail, we update $\mathbf{W}^{*}_{2}$ by using the Hebbian's rule:
\begin{equation}\label{updatew2}
    \mathbf{W}^{*}_{2} = \mathbf{W}_2 +  \varepsilon \big( \mathbf{\hat{A}}_{v} \mathbf{X}_{v} \mathbf{W}_1 \big)^{T} \mathbf{H}_{v-1},
\end{equation}
where $\varepsilon$ refers to the learning rate. The latter item is a detailed expansion of Eq.~\eqref{deltaW}.
$\mathbf{W}^{*}_{2}$ emphasizes the consistent information between old and new representations. 
Herein, $\mathbf{W}^{*}_{2}$ reconstruct the weight through Hebbian's rule, which is more suitable for the intermingling of old and new knowledge.
Then, it acts on forward propagation to acquire new representation and can be denoted as,
\begin{equation}\label{Hvfinal}
    \mathbf{H}^{*}_{v} = \mathbf{H}_v + \varepsilon \big( \mathbf{\hat{A}}_{v} \mathbf{X}_{v} \mathbf{W}_1 \big) \mathbf{W}^*_2,
\end{equation}
where $\mathbf{H}^{*}_{v} \in \mathbb{R}^{n\times c}$ is the final fused representation after structured reinforcement. 
By mining the inherent structure in the multi-view data, structured Hebbian plasticity enables more efficient learning, and improved domain generalization. 

\subsection{Synaptic Partition Learning}
In the process of biological neural networks, emphasizing only relevant synaptic connections without suppressing irrelevant synaptic connections may result in the model being interfered with by a large amount of noise and redundant information.
Thus, we design a synaptic partition learning mechanism, which focuses on partitioning synapses into distinct groups or partitions and independently updating the synaptic weights within each partition.  
We adopt it to enable more efficient and targeted learning by selectively updating subsets of synapses.

Specifically, we first define a one-dimensional matrix of the same size as the weight $\mathbf{W}_1$ and mask it randomly, ensuring that the number of places 1 satisfies a given ratio $\theta$, as shown below.
\begin{equation}
    \mathbf{M} = \mathbf{M} \cdot \mathbf{S},
\end{equation}
where $\mathbf{M} \in \mathbb{R}^{nd \times 1}$ is an all-1 matrix and $\mathbf{S} \in \mathbb{R}^{nd \times 1}$ a randomly initialized matrix for each epoch. 
Here, the number of elements in $\mathbf{S}$ that have 1 value is no more than $\lfloor \theta \cdot \mathbf{M}.size() \rfloor$.
We then partially disable the neurons with randomly initialized $\mathbf{M}$.
As we can see, $\mathbf{W}_1$ can be represented as,
\begin{equation}\label{mask}
    \mathbf{W}^{*}_1 = \mathrm{Reshape}(\mathbf{M}) \cdot \mathbf{W}_1.
\end{equation}
$\mathrm{Reshape}(\mathbf{M}) \in \mathbb{R}^{d_v \times d}$ is a function to adjust the dimension of the weight $\mathbf{W}_1$.
In each round of training, we set $\theta \ll 1/V$  and use $\theta$ to randomize the setup of $\mathbf{S}$.
Instead of globally adjusting all synapses simultaneously, synaptic partition learning allows for local adjustments within specific subsets of synapses, potentially leading to faster convergence, improved generalization, and reduced susceptibility to oblivion of knowledge.

In summary, the synaptic partitioning learning module aims to eliminate irrelevant connections by inhibiting some synapses, thus reducing the interference of redundant information.
While the structured Hebbian plasticity module focuses on activating highly correlated synapses to facilitate meaningful information transfer. 
The synergistic effect of these two modules can improve the generalization ability of the model.

\subsection{Loss Function}
Moreover, our model attempts to regulate the entire activity level of neurons to keep it within a certain range. 
Thus, we adopt a loss function to penalize shifting weights and thus mitigate large shifts in weights.
If the activity of certain neurons begins to change excessively, this loss works to restore balance, i.e, 
\begin{equation}\label{f2loss}
	\mathcal{L}_{RE} = \frac{1}{2} \sum_{i=1}^{2} \| \mathbf{W}_{i} - \big(\mathbf{W}_{i}\big)_{old} \| _{F}^{2},
\end{equation}
where $\big(\mathbf{W}_{1}\big)_{old}$ and $\big(\mathbf{W}_{2}\big)_{old}$ are the weights of the previous representation.
We expect the two shared weights $\mathbf{W}_{1}$ and $\mathbf{W}_{2}$ not to fluctuate unduly, thus maintaining the holistic memory of the old knowledge.

Besides, for semi-supervised node classification, we calculate CrossEntropy loss and update parameters in our method, as follows
\begin{equation}\label{lceloss}
	\mathcal{L}_{CE} = - \sum_{i \in \Omega} \sum_{j=1}^{c} {y}_{ij}  \ln\hat{y}_{ij},
\end{equation}
where $\Omega$ refers to the set of labeled samples, $\hat{y} = \mathrm{softmax} (\mathbf{H}^{*}_{v})$ and $c$ is the amount of classes.
The final loss can be expressed as
\begin{equation}\label{finalloss}
	\mathcal{L}_{all} = \mathcal{L}_{CE} + \beta \mathcal{L}_{RE}.
\end{equation}
$\mathcal{L}_{all}$ is the final loss for our method, which consists of two components.
$\beta$ is a parameter that resizes the loss $\mathcal{L}_{RE}$.
We conclude the training procedure of MVIL in Algorithm~\ref{algotirhmMVIL}. 
\begin{algorithm}[h]
	\caption{MVIL}
	\label{algotirhmMVIL}
	\begin{algorithmic}
		\REQUIRE{Multi-view feature matrices $\mathbf{\{X\}}_{v=1}^{V}$, label matrix $\mathbf{Y} \in \mathbb{R}^{|\Omega| \times c}$, hyperparameter $k$ in $k$NN.  }
		\ENSURE{Predictive output $\mathbf{H}^{*}_{v}$. }
        \FOR{new view data $\mathbf{X}_{v}$ comes...}
		\WHILE{not convergent}
		\STATE { Construct the adjacency $\mathbf{A}_v$ by utilizing $k$NN;}
        \STATE{ Calculate the weight $\mathbf{W}_{1}^{*}$ by Equation \eqref{mask};}
		\STATE{ Calculate $\mathbf{H}_{v}$ by Equation \eqref{Hv_initial};}
        \STATE{ Update the $\mathbf{W}_2^{*}$ by Equation \eqref{updatew2};}
		\STATE{ Obtain the final representation $\mathbf{H}_{v}^{*}$ by Equation \eqref{Hvfinal}; }
		\STATE{ Compute the loss $\mathcal{L}_{RE}$ and $\mathcal{L}_{CE}$ by Equations \eqref{f2loss} and \eqref{lceloss};}
		\STATE{ Optimize the trainable parameters from network by backward propagation;}
		\ENDWHILE
        \ENDFOR
		\STATE{ Obtain the predictive output $\mathbf{H}_{v}^{*}$; }
		\RETURN{ Predictive output $\mathbf{H}_{v}^{*}$. }
	\end{algorithmic}
\end{algorithm}

\section{Experiments}
To further validate the effectiveness of the proposed method, we have designed the experimental section intended to answer the following key evaluation questions (EQs):
\begin{itemize}
	\item \textbf{EQ1} Does MVIL achieve superior performance compared to its competitors for the semi-supervised node classification task?
	\item \textbf{EQ2} Does MVIL capture the tight association between old and new views when new ones add up, and does it promote learning about the whole knowledge?
\end{itemize}
MVIL has performed ablation experiments, parameter sensitivity analysis, and t-Sne demonstration in the Appendix.
\subsection{Datasets \& Compared Methods}
\begin{table}[!ht]
\resizebox{\linewidth}{!}{
    \centering
    
    \begin{tabular}{cccccc}
    \midrule
        Datasets & Views & Samples & Features & Classes & Data Types \\ \midrule
        100leaves & 3 & 1,600 & 64 & 100 & plant species \\ 
        Animals & 2 & 10,158 & 4,096 & 50 & images \\ 
        Flower17 & 7 & 1,360 & 7 & 17 & images \\ 
        NGs & 3 & 500 & 2,000 & 5 & newsgroup document \\ 
        NoisyMNIST\_15000 & 2 & 15,000 & 784 & 10 & images \\ 
        YaleB\_Extended & 5 & 2,424 & 1,024 & 38 & images \\ \midrule
    \end{tabular}}
    \caption{A brief description of the test multi-view datasets.}\label{DataDescription}
\end{table}
\subsubsection{Datasets}
We adopt six multi-view graph datasets widely used in different domains to evaluate the performance of MVIL compared to state-of-the-art baselines.
Table \ref{DataDescription} illustrates the details of all six datasets.

\begin{table*}[]
\centering
    \resizebox{0.95\textwidth}{!}{
\begin{tabular}{cc|ccc|ccccc}
\midrule
\multicolumn{2}{c|}{Classification}                              & \multicolumn{3}{c|}{Static Multi-view Learning}                            & \multicolumn{5}{c}{Multi-view Incremental Learning}                                                                                         \\ \midrule
\multicolumn{1}{c|}{Datasets}                           & Metric & TMC                   & LGCNFF       & IHGCN                               & GAT          & SI           & MAS          & \multicolumn{1}{c|}{MVCIL}                               & MVIL                                \\ \midrule
\multicolumn{1}{c|}{\multirow{4}{*}{100leaves}}         & ACC    & 79.20 (0.08)          & 87.41 (0.62) & \underline{91.03 (0.70)} & 43.52 (0.94) & 43.25 (0.25) & 43.46 (0.32) & \multicolumn{1}{c|}{69.50 (0.43)}                        & \textbf{91.50 (0.47)}               \\
\multicolumn{1}{c|}{}                                   & P      & 79.20 (0.08)          & 88.76 (0.83) & \underline{92.29 (0.53)} & 42.36 (1.64) & 45.16 (1.18) & 45.73 (1.30) & \multicolumn{1}{c|}{70.31 (0.06)}                        & \textbf{92.55 (0.46)}               \\
\multicolumn{1}{c|}{}                                   & R      & 80.56 (0.07)          & 87.41 (0.62) & \underline{91.03 (0.70)} & 43.52 (0.94) & 43.25 (0.25) & 43.46 (0.32) & \multicolumn{1}{c|}{69.50 (0.43)}                        & \textbf{91.50 (0.47)}               \\
\multicolumn{1}{c|}{}                                   & MAF1   & 78.50 (0.09)          & 86.58 (0.56) & \underline{90.76 (0.80)} & 40.27 (0.96) & 40.58 (0.24) & 40.89 (0.24) & \multicolumn{1}{c|}{67.87 (0.30)}                        & \textbf{91.33 (0.48)}               \\ \midrule
\multicolumn{1}{c|}{\multirow{4}{*}{Animals}}           & ACC    & 81.35 (0.53)          & 74.15 (1.32) & \underline{83.13 (0.04)} & 39.20 (0.36) & 45.86 (0.14) & 46.29 (0.16) & \multicolumn{1}{c|}{67.03 (0.03)}                        & \textbf{84.42 (0.10)}               \\
\multicolumn{1}{c|}{}                                   & P      & 74.23 (0.65)          & 69.07 (1.71) & \underline{80.46 (0.35)} & 34.18 (1.33) & 39.92 (0.18) & 39.89 (0.28) & \multicolumn{1}{c|}{64.86 (0.10)}                        & \textbf{81.59 (0.31)}               \\
\multicolumn{1}{c|}{}                                   & R      & \textbf{79.19 (1.01)} & 65.49 (1.46) & 76.17 (0.04)                        & 32.45 (0.44) & 38.15 (0.04) & 38.59 (0.18) & \multicolumn{1}{c|}{60.53 (0.03)}                        & \underline{78.10 (0.19)} \\
\multicolumn{1}{c|}{}                                   & MAF1   & 74.65 (0.76)          & 65.20 (1.59) & \underline{76.82 (0.03)} & 31.53 (0.69) & 37.03 (0.13) & 37.19 (0.02) & \multicolumn{1}{c|}{61.05 (0.00)}                        & \textbf{78.32 (0.14)}               \\ \midrule
\multicolumn{1}{c|}{\multirow{4}{*}{Flower17}}          & ACC    & 52.44 (0.63)          & 45.35 (4.20) & \underline{58.94 (0.17)} & 17.78 (1.56) & 44.81 (0.20) & 47.10 (0.20) & \multicolumn{1}{c|}{40.93 (1.63)}                        & \textbf{60.54 (0.72)}               \\
\multicolumn{1}{c|}{}                                   & P      & 52.44 (0.63)          & 55.44 (6.80) & \underline{62.43 (0.27)} & 10.53 (3.75) & 43.26 (0.22) & 45.82 (0.33) & \multicolumn{1}{c|}{42.01 (1.62)}                        & \textbf{61.98 (1.01)}               \\
\multicolumn{1}{c|}{}                                   & R      & 54.06 (0.48)          & 45.32 (4.20) & \underline{58.94 (0.17)} & 17.78 (1.56) & 44.81 (0.20) & 47.10 (0.20) & \multicolumn{1}{c|}{40.93 (1.63)}                        & \textbf{60.54 (0.72)}               \\
\multicolumn{1}{c|}{}                                   & MAF1   & 50.16 (0.74)          & 39.75 (5.54) & \underline{59.14 (0.20)} & 10.19 (2.77) & 42.86 (0.22) & 45.61 (0.31) & \multicolumn{1}{c|}{38.53 (2.18)}                        & \textbf{60.22 (0.70)}               \\ \midrule
\multicolumn{1}{c|}{\multirow{4}{*}{NGs}}               & ACC    & 93.11 (0.01)          & 90.65 (1.27) & \textbf{96.89 (0.14)}               & 72.52 (4.74) & 79.11 (1.33) & 83.56 (0.67) & \multicolumn{1}{c|}{58.89 (7.11)}                        & \underline{95.04 (0.64)} \\
\multicolumn{1}{c|}{}                                   & P      & 93.11 (0.01)          & 92.02 (0.66) & \textbf{96.89 (0.14)}               & 75.00 (4.66) & 82.56 (0.07) & 84.13 (0.62) & \multicolumn{1}{c|}{60.38 (8.56)}                        & \underline{95.16 (0.60)} \\
\multicolumn{1}{c|}{}                                   & R      & 93.14 (0.01)          & 90.67 (1.27) & \textbf{96.89 (0.14)}               & 72.52 (4.74) & 79.11 (1.33) & 83.56 (0.67) & \multicolumn{1}{c|}{58.89 (7.11)}                        & \underline{95.04 (0.64)} \\
\multicolumn{1}{c|}{}                                   & MAF1   & 93.03 (0.01)          & 90.57 (1.23) & \textbf{96.87 (0.14)}               & 72.50 (5.12) & 79.31 (1.30) & 83.60 (0.64) & \multicolumn{1}{c|}{58.81 (7.67)}                        & \underline{95.05 (0.64)} \\ \midrule
\multicolumn{1}{c|}{\multirow{4}{*}{NoisyMNIST\_15000}} & ACC    & 87.01 (0.35)          & 89.78 (0.48) & \textbf{93.64 (0.46)}               & 73.43 (0.61) & 82.74 (2.10) & 90.02 (0.29) & \multicolumn{1}{c|}{90.79 (0.47)}                          & \underline{92.43 (0.64)} \\
\multicolumn{1}{c|}{}                                   & P      & 87.03 (0.26)          & 89.75 (0.47) & \textbf{93.61 (0.47)}               & 74.02 (0.87) & 85.44 (0.97) & 90.20 (0.04) & \multicolumn{1}{c|}{90.64 (0.46)}                          & \underline{92.55 (0.49)} \\
\multicolumn{1}{c|}{}                                   & R      & 86.59 (0.37)          & 89.58 (0.49) & \textbf{93.51 (0.48)}               & 72.89 (0.63) & 82.45 (2.07) & 89.80 (0.30) & \multicolumn{1}{c|}{90.60 (0.47)}                          & \underline{92.29 (0.62)} \\
\multicolumn{1}{c|}{}                                   & MAF1   & 86.52 (0.40)          & 89.58 (0.49) & \textbf{93.53 (0.47)}               & 72.43 (0.71) & 82.25 (1.96) & 89.83 (0.28) & \multicolumn{1}{c|}{90.59 (0.48)}                          & \underline{92.30 (0.62)} \\ \midrule
\multicolumn{1}{c|}{\multirow{4}{*}{YaleB\_Extended}}   & ACC    & 50.59 (1.10)          & 34.01 (1.43) & 47.69 (0.36)                        & 31.19 (0.82) & 32.76 (0.25) & 32.86 (0.16) & \multicolumn{1}{c|}{\underline{64.16 (0.82)}} & \textbf{66.99 (0.28)}               \\
\multicolumn{1}{c|}{}                                   & P      & \textbf{81.62 (0.84)} & 48.32 (6.29) & \underline{75.85 (0.17)} & 60.14 (4.80) & 36.12 (0.33) & 36.98 (0.45) & \multicolumn{1}{c|}{67.45 (0.97)}                        & 71.06 (0.23)                        \\
\multicolumn{1}{c|}{}                                   & R      & 50.61 (1.10)          & 33.99 (1.40) & 47.69 (0.36)                        & 31.21 (0.86) & 32.83 (0.25) & 32.92 (0.15) & \multicolumn{1}{c|}{\underline{64.24 (0.82)}} & \textbf{67.07 (0.28)}               \\
\multicolumn{1}{c|}{}                                   & MAF1   & 58.13 (0.77)          & 33.24 (3.38) & 53.78 (0.26)                        & 34.01 (1.87) & 33.51 (0.23) & 33.85 (0.00) & \multicolumn{1}{c|}{\underline{64.86 (0.81)}} & \textbf{67.70 (0.27)}               \\ \midrule
\end{tabular}}\caption{Node classification performance with various algorithms.}\label{Performence}
\end{table*}
\subsubsection{Compared Methods}
We compare the proposed method with classical and state-of-the-art baseline models, including GAT~\cite{velivckovic2018graph}, SI~\cite{zenke2017continual}, MAS~\cite{aljundi2018memory}, LGCNFF~\cite{chen2023learnable},  IHGCN~\cite{zou2024revisiting}, TMC~\cite{han2020trusted} and MVCIL~\cite{li2024multi}.

\subsection{Experimental Setting}

Experimental rounds of the proposed method are 600 per view under 10\% randomly labeled data and the average value is taken by repeating the run 3 times. 
Four well-known classification evaluation metrics are applied to our experiments, including Accuracy (ACC), Precision (P), Recall (R), and Macro-averaged F1 Score (MAF1). 
ACC is defined as the number of correctly classified samples divided by the number of samples.
P reflects how accurate the model is in predicting the positive class while R is the ability to recognize samples in the positive class. MAF1 is the arithmetic mean of the F1 values over all classes.
The values of these metrics fall within the range of [0,1], where higher values indicate better performance.

\subsection{Comparison to SOTA (EQ1)}
We next evaluate the effectiveness of our method on the node classification task compared with several classical and state-of-the-art methods in Table~\ref{Performence}, where the best performance is highlighted in bold and the second-best results are underlined.
We divide the compared methods into two classes: static multi-view learning and view incremental learning.
\begin{figure}[!t]
	\centering
	\includegraphics[width=0.9\linewidth]{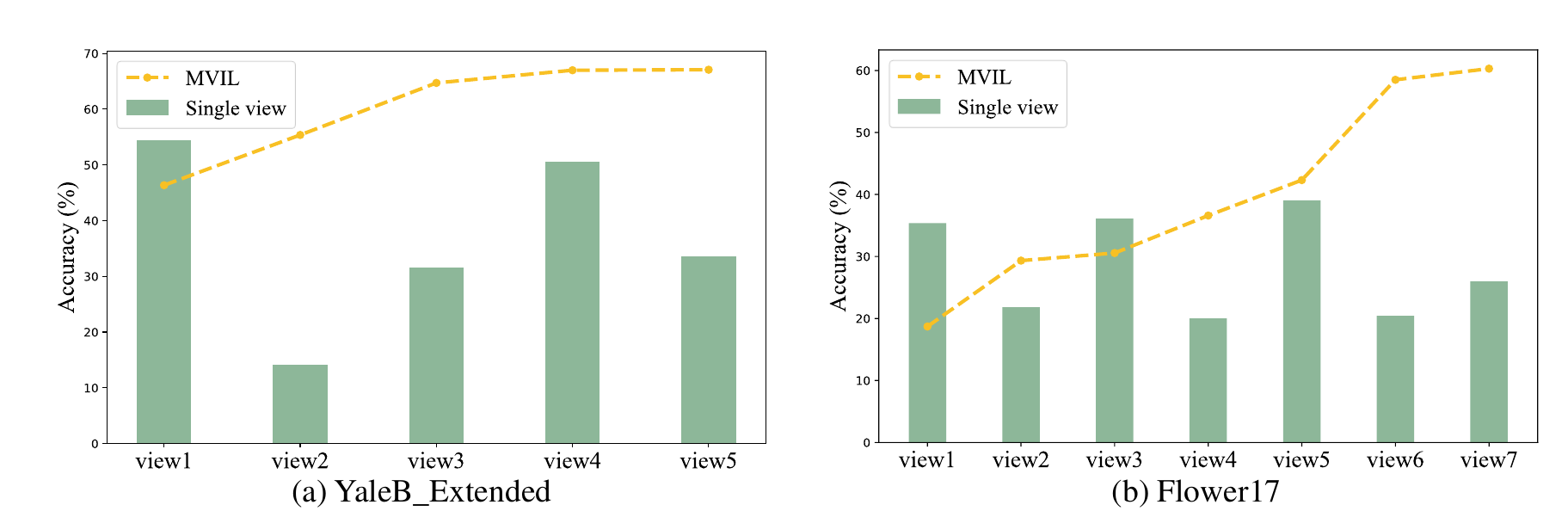}\\
	\caption{A comparison of MVIL streaming input and single-view learning performance.}
	\label{zx}
\end{figure}
\subsubsection{Static Multi-view Learning}
We compare it with the leading multi-view methods, where all view information is collected completely and then trained on the left side of Table \ref{Performence}.
The three compared methods, TMC, LGCNFF, and IHGCN simultaneously obtain all the views and can be well mined for correlations between the views. 
TMC skillfully blends credible evidence of view judgment by determining the confidence level of view.
In contrast, LGCNFF and IHGCN achieve very impressive performance by capturing consistent and complementary information of multi-view data.
In particular, static multi-view learning obtains leading performance in datasets with a small number of views and excellent view quality; in the face of datasets with a large number of views, it is very challenging to extract the overall cognition and determine the view quality.
\subsubsection{Multi-view Incremental Learning}
First, we introduce GAT into the view incremental framework and we find that GAT is explicitly failing to memorize the old view information. 
Their performance is intricately linked to the quality of the most recent view data rather than the integration of the overall information.
\begin{figure}[!t]
	\centering
	\includegraphics[width=0.9\linewidth]{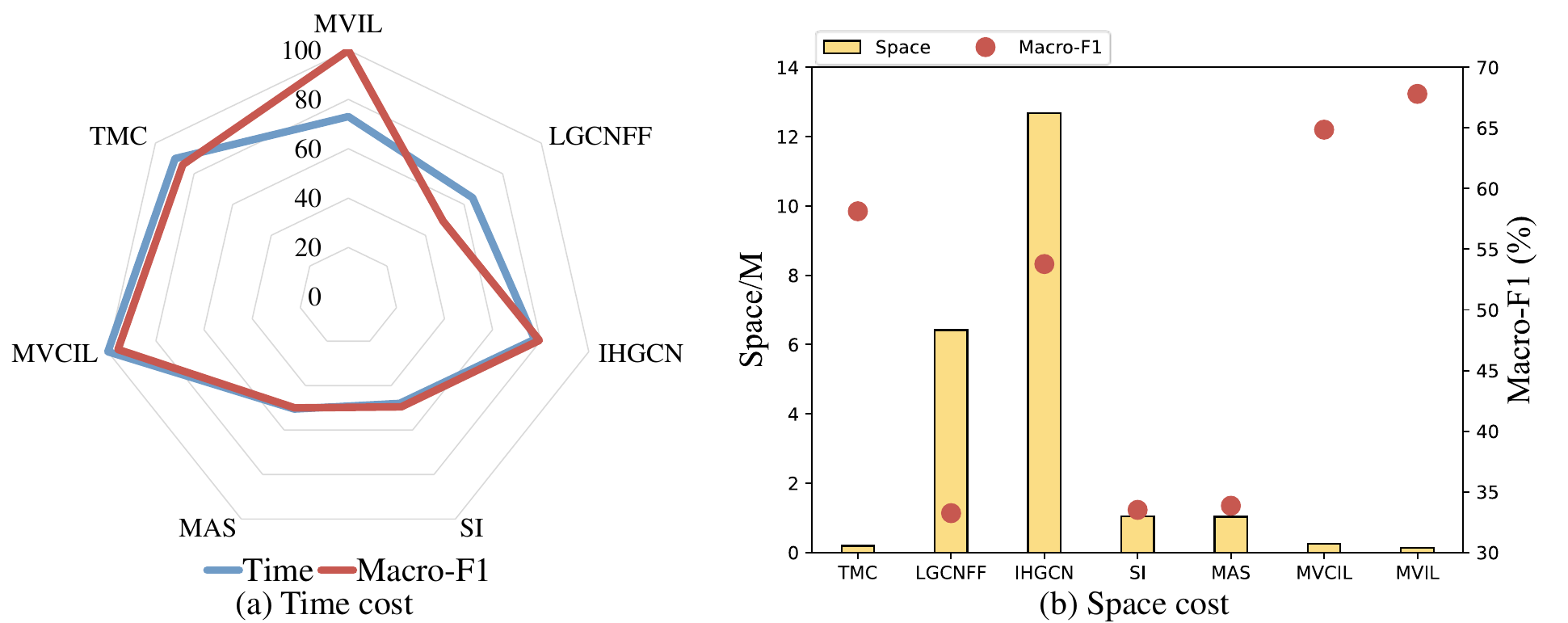}\\
	\caption{The time-space cost of all compared algorithms.}
	\label{time}
\end{figure}
We then compare this to the continual learning approaches associated with synaptic plasticity.
SI and MAS express the role of synaptic plasticity in capturing overall information, but it is not easy to reconcile the poor quality of views and the large number of categories.
Moreover, we compare it with relevant methods for multi-view class-incremental learning MVCIL. MVCIL shows excellent performance in learning datasets with multiple view counts due to its ability to memorize old knowledge well.

From the Table~\ref{Performence}, it is evident that the proposed method has achieved a significant leading position. 
More specifically, MVIL improves over the suboptimal method by \textbf{1.6\%} in the Flower17 dataset and \textbf{2.83\%} in the YaleB\_Extended dataset.
This illustrates that even with non-static multi-view data, we are able to capture a comprehensive representation of them as the number of views increases.
Incidentally, it reaffirms the ability of our model to extract maximum information about the whole view and fuse the ever-growing view data in a comprehensive and efficient manner.
\begin{figure}[!t]
	\centering
	\includegraphics[width=0.85\linewidth]{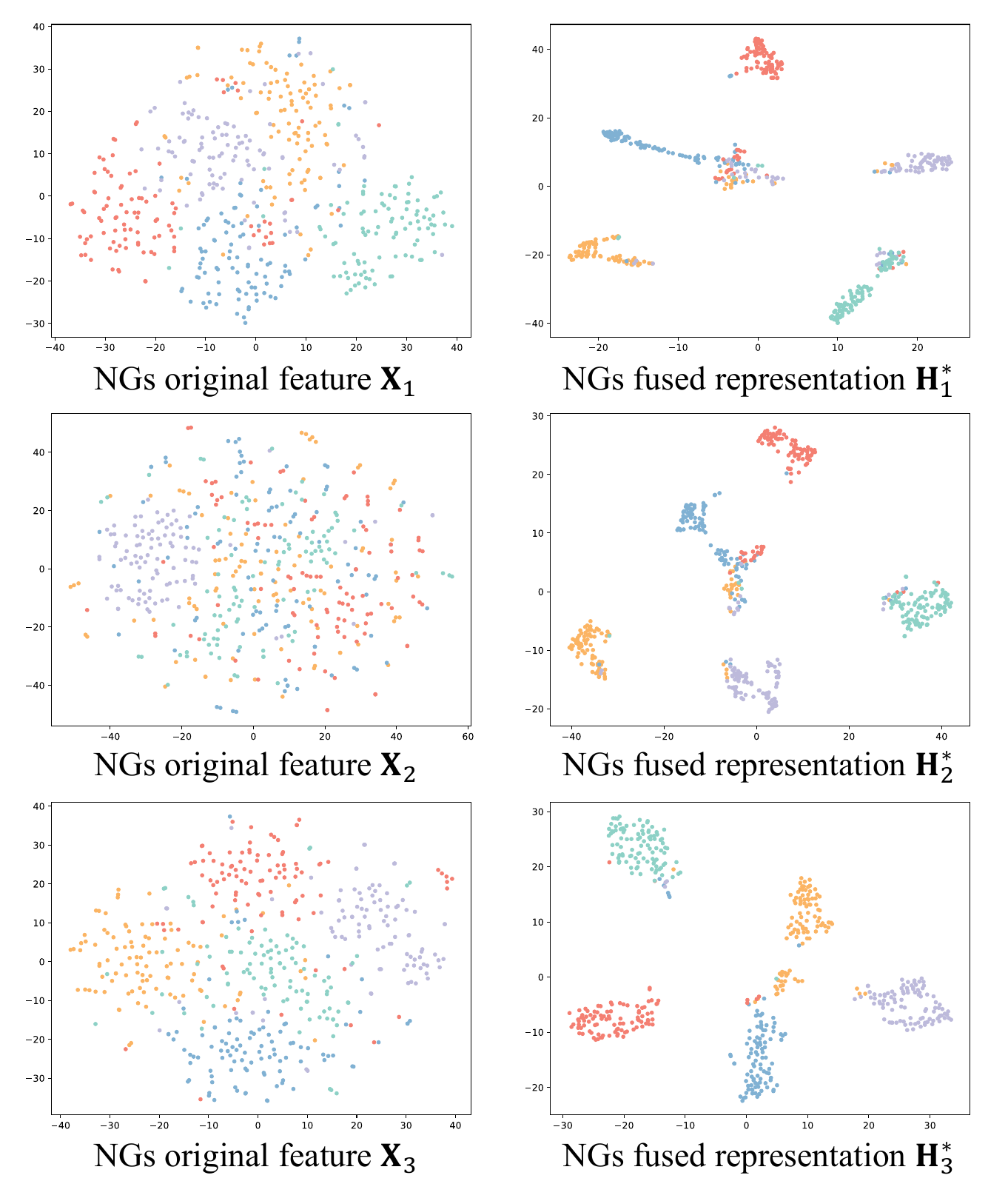}\\
	\caption{The initial features and learned fused representations for each view of the NGs.}
	\label{NGs}
\end{figure}
\subsubsection{Time and Space Cost}

To validate the low energy consumption and high efficiency of MVIL, we plotted the time-space cost shown in Fig.~\ref{time}. 
Herein, Fig. 4(a) illustrates all methods' time (as a percentage of the longest time) and performance (as a percentage of the highest performance), while Fig. 4(b) depicts space cost and performance. 
Our MVIL utilizes very few parameters and incurs minimal time costs while delivering leading performance.
\subsection{Knowledge Accumulation (EQ2-1)}
We plot the change in overall model performance as the views continue to increase, and add the results of training with only one view as a comparison, on Fig.~\ref{zx}. 
It is evident that the performance of our MVIL improves as the number of views increases, surpassing the performance achieved with single-view learning. However, partial single-view learning sometimes outperforms our MVIL. 
This is because our synaptic partitioning learning module imposes restrictions on neuronal activity, which, while aiding in the retention of old knowledge, can also limit performance. 

This not only confirms the effectiveness of the multi-view incremental strategy but also clearly indicates its ability to outperform traditional single-view learning paradigms. 
Specifically, by fusing the complementary information embedded in multiple views, the model is able to understand the data features more comprehensively, which in turn leads to more accurate and robust prediction or classification. 
This process not only validates our model's superior ability to accumulate and integrate knowledge but also demonstrates its ability to retain old knowledge while continuously absorbing and utilizing new knowledge from additional views.

\subsection{Knowledge Leads to Learning (EQ2-2)}
To further validate the ability of our MVIL to integrate information from all views, we plot the initial features and the learned fused representation for each new view data, as in Fig.~\ref{NGs}.
In the original feature $\mathbf{X}_1$, the inter-class and intra-class distances of the initial features are almost equal, yet it is fortunate that the intra-class nodes are clustered together. 
Thus $\mathbf{H}^{*}_1$ is well differentiated. 
When view data $\mathbf{X}_2$ comes in, the features are muddled and the same class of nodes is scattered, but because of the favorable guiding effect of $\mathbf{H}^{*}_1$, the inter-class distance of $\mathbf{H}^{*}_2$ is still large, but the intra-class nodes become slightly decentralized.
When view data $\mathbf{X}_3$ enters, the features are sharply differentiated, and with the guiding effect of $\mathbf{H}^{*}_2$, the nodes in the middle of the two classes are pulled back into the class.

Due to the consideration of fine-grained correlations between nodes, the old representations can still guide the representation learning well even if the new view data is scattered, and the learned representations adaptively fuse the merits of the two. 
This adequately illustrates the importance of structured Hebbian plasticity. 
When the new incoming view data is complementary to the old knowledge, it learns a well orthogonal representation to accommodate both; when the new data is not well differentiated, the module takes full advantage of the guidance of the old view in order to learn serviceable representations from it.
We capitalize on the association between the old and new views, thus mitigating the effects of poor view quality on a single view.
It not only enhances the accuracy in understanding the overall view but continuously improves MVIL's perceptual effectiveness.
\section{Conclusion}
Traditional multi-view learning paradigms are constrained by static multi-view data, impeding their ability to effectively accommodate the dynamic and ever-expanding amount of view data encountered in real-world settings. 
In this study, drawing inspiration from biological neurology, we proposed a multi-view incremental learning framework named MVIL. 
Therein, the underlying problem structure remains the same, but the input distribution changes. 
Different from continual learning, multi-view incremental learning prioritizes overall cognition when interacting with streaming data and each task includes only one view.
Our method first applied Hebbian learning principles to multi-view graph representation learning.
We devised a weight adjustment strategy that includes structured Hebbian plasticity and synaptic partitioning learning, which fine-grain fused representations thereby better mimicking the brain's cognitive decision-making.
Experimental evaluations demonstrate the ability of the proposed MVIL to effectively fine-grained fusion of ever-increasing view representations.

\section{Acknowledgments}
This work is in part supported by the National Key Research and Development Plan of China under Grant 2021YFB3600503 and National Natural Science Foundation of China under Grants No.U21A20472, No.62276065, No.62204140, No.62476035 and No.62206037.

\bibliography{aaai25}
\clearpage

\appendix

\section{Knowledge Accumulation}
\begin{figure}[!h]
	\centering
	\includegraphics[width=\linewidth]{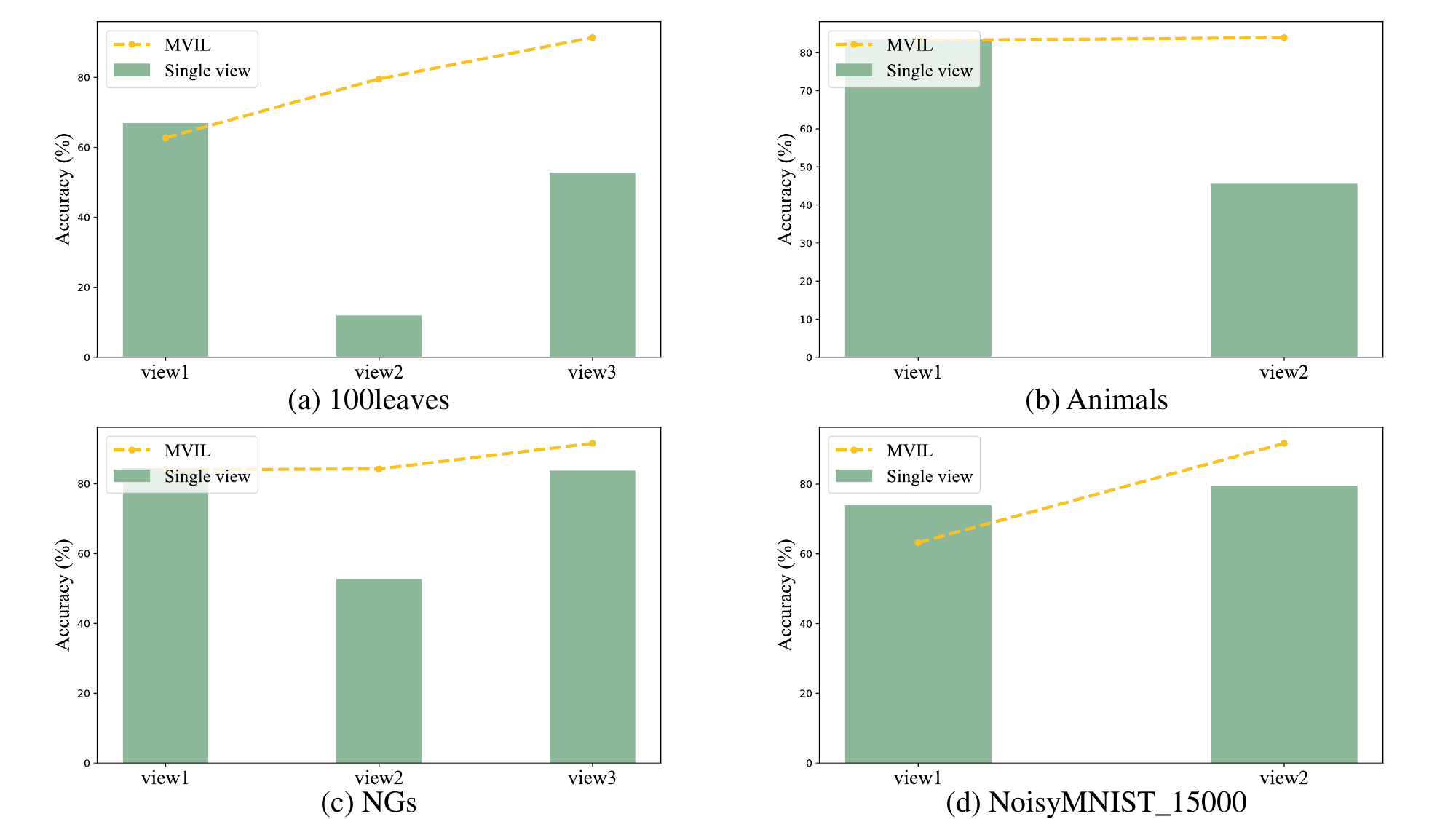}\\
	\caption{A comparison of MVIL streaming input and single-view learning performance.}
	\label{zx}
\end{figure}
We plot the change in overall model performance as the number of views increases and include the results of training with only one view for comparison, as shown in Fig.~\ref{zx}. 
We observe that in datasets with a limited number of views, low-quality views can significantly impact overall recognition. 
However, our MVIL effectively guides the learning from these low-quality perspectives by leveraging prior knowledge to achieve a more comprehensive understanding.

This process not only confirms our model's exceptional ability to accumulate and integrate knowledge, but highlights its proficiency in retaining previously information while continuously absorbing and leveraging new insights from additional views. 
This ability to seamlessly integrate and build upon existing knowledge underscores the model's overall strength and adaptability in dynamic environments.

\section{Ablation Study}
In this section, we perform ablation study to verify the validity of the proposed module in Table~\ref{ablatable}, where the best performance is highlighted in bold and the second-best results are underlined.
We divide the main body of the method into three components, namely the synaptic partition learning mechanism $C_1$, weights reconstructed loss $C_2$, and structured Hebbian plasticity $C_3$.
It is obvious to observe that $C_1$ plays a great role for the dataset with a small number of views, but the performance improvement is not high for the two datasets with a large number of views, flower17 and YaleB\_Extended. 
We hypothesize that this is due to the high number of masked synapses.
And $C_2$ is effective for performance retention, especially in datasets with a high number of views, it can retain old knowledge well. 
Therefore, the performance of fusing $C_1$ and $C_2$ can be adapted to datasets with various numbers of views. 
In addition, adding $C_3$ enables a fine-grained weight structure construction, which improves the performance quite a lot.
\begin{table}[!h]
\resizebox{\linewidth}{!}{
\begin{tabular}{c|c|cccc}
\midrule
Datasets                           & Metric & $C_1$           & $C_2$           & $C_1$+$C_2$        & $C_1$+$C_2$+$C_3$          \\\midrule\midrule
\multirow{2}{*}{100leaves}         & ACC    & 90.67 (0.17) & 86.79 (0.20) & \underline{90.79 (0.69)} & \textbf{91.36 (0.68)} \\
                                   & MAF1   & 90.44 (0.20) & 86.59 (0.23) & \underline{90.52 (0.69)} & \textbf{91.36 (0.68)} \\\midrule
\multirow{2}{*}{Animals}           & ACC    & 79.97 (0.06) & 80.00 (0.05) &  \underline{84.01 (0.11)} & \textbf{84.42 (0.10)} \\
                                   & MAF1   & 69.81 (0.07) & 69.86 (0.00) &  \underline{77.33 (0.10)} & \textbf{78.32 (0.14)} \\\midrule
\multirow{2}{*}{Flower17}          & ACC    & 36.00 (0.21) & 53.65 (0.25) &  \underline{56.21 (0.42)} & \textbf{57.79 (0.37)} \\
                                   & MAF1   & 34.29 (0.23) & 53.30 (0.27) &  \underline{55.94 (0.34)} & \textbf{57.42 (0.37)} \\\midrule
\multirow{2}{*}{NGs}               & ACC    & 83.33 (0.65) & 83.70 (0.10) &  \underline{94.37 (0.42)} & \textbf{95.04 (0.64)} \\
                                   & MAF1   & 83.37 (0.66) & 83.70 (0.10) &  \underline{94.37 (0.42)} & \textbf{95.05 (0.64)} \\\midrule
\multirow{2}{*}{NoisyMNIST\_15000} & ACC    & \textbf{95.05 (0.64)} & 91.38 (0.11) & 91.91 (0.12) &  \underline{92.43 (0.64)} \\
                                   & MAF1   & 89.74 (0.15) & 91.23 (0.11) &  \underline{91.74 (0.13)} & \textbf{92.30 (0.62)} \\\midrule
\multirow{2}{*}{YaleB\_Extended}   & ACC    & 34.70 (0.58) &  \underline{66.62 (0.07)} & 62.66 (0.16) & \textbf{66.99 (0.28)} \\
                                   & MAF1   & 35.10 (0.52) &  \underline{67.31 (0.05)} & 63.08 (0.17) & \textbf{67.70 (0.27)}\\\midrule
\end{tabular}}
\caption{Ablation experiments for different modules.}
	\label{ablatable}
\end{table}

\begin{figure}[!h]
	\includegraphics[width=\linewidth]{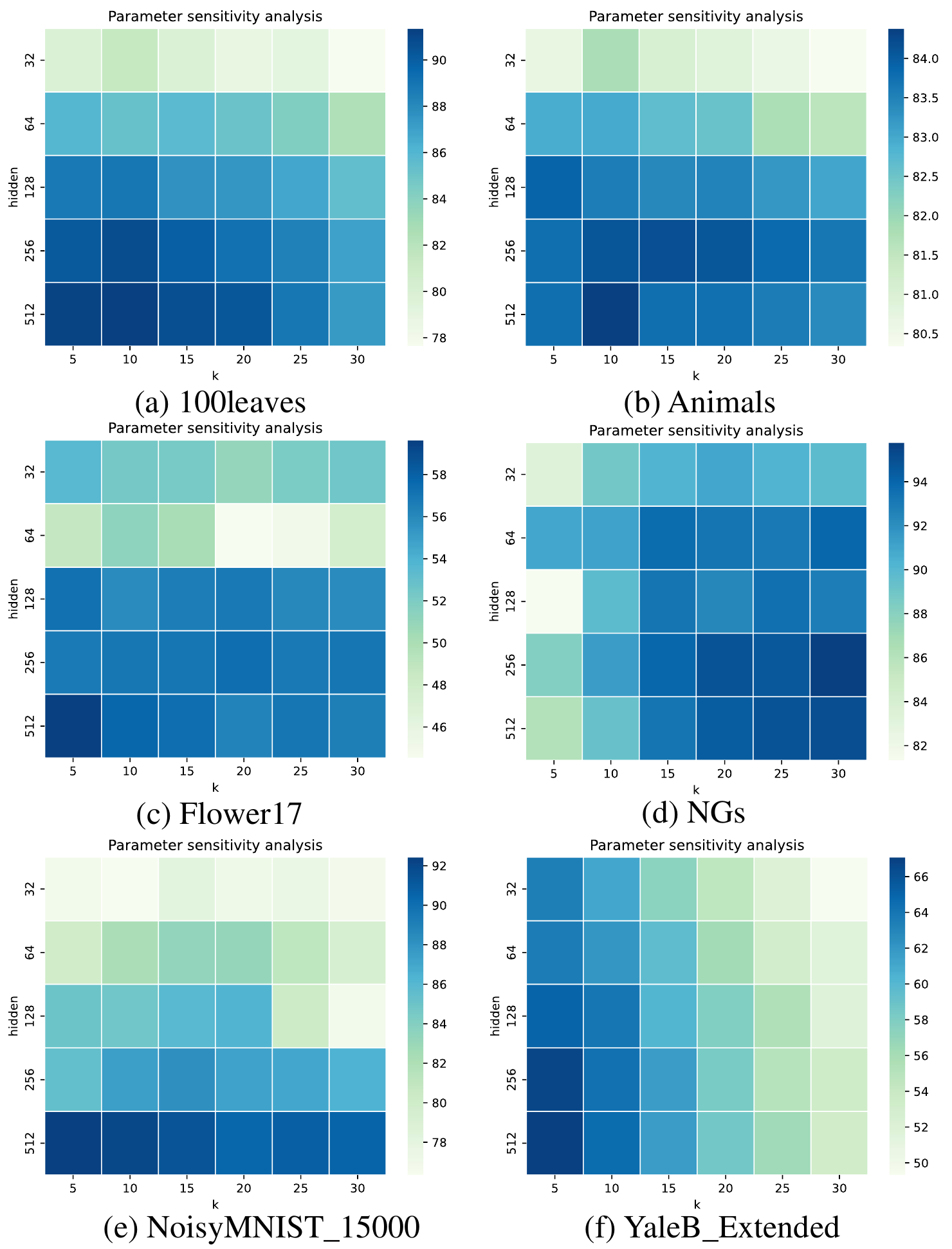}\\
	\caption{The parameter sensitivity analysis of $k$ and hidden in MVIL on tested datasets.}
	\label{pa}
\end{figure}

\begin{figure*}[!ht]
	\centering
	\includegraphics[width=\textwidth]{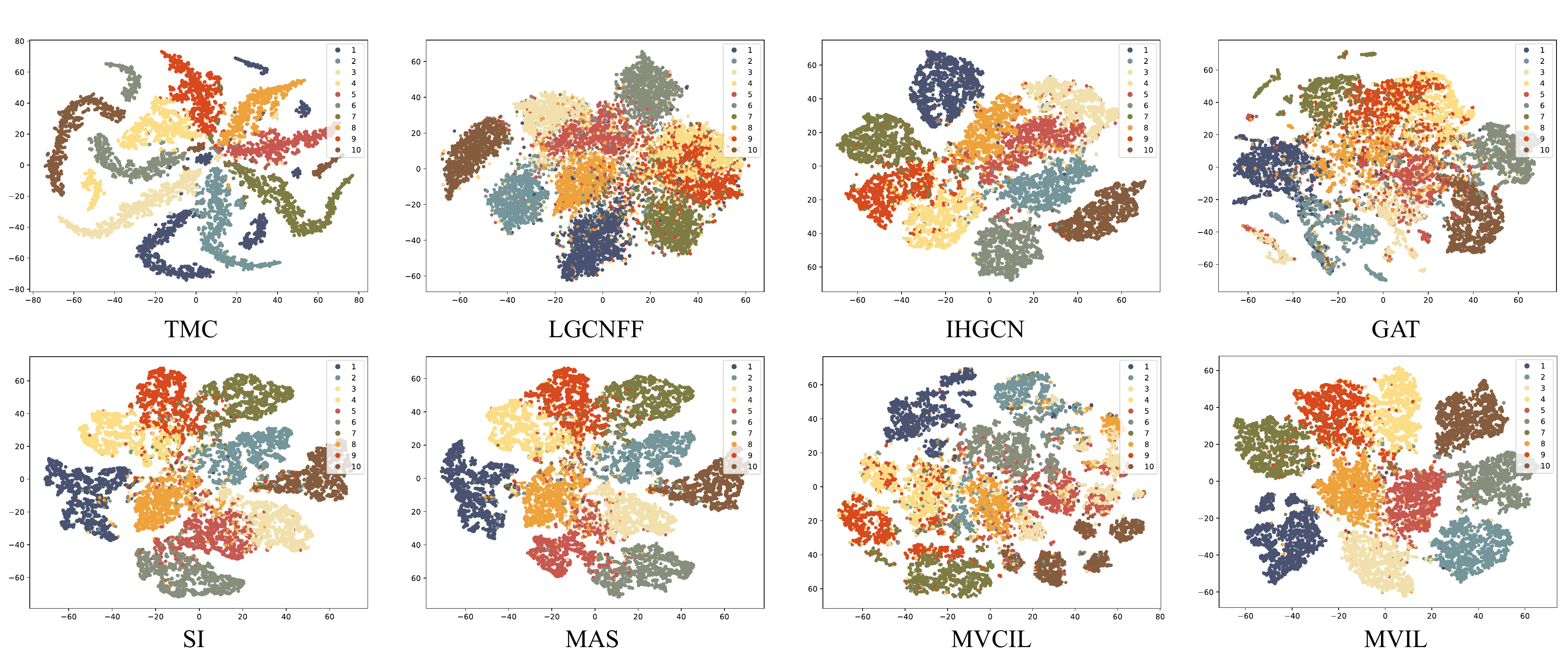}\\
	\caption{The visualization for multi-view semi-supervised classification on NoisyMNIST\_15000.}
	\label{Tsne}
\end{figure*}

\section{Parameter Settings}
Table~\ref{ParaDescription} presents the list of parameters utilized for the dataset in the experiments.
Here, $lr$ denotes the learning rate, hidden layer $d$ refers to the GCN middle layer's network parameters, and $\beta$ represents the scaling value in loss $\mathcal{L}_{RE}$.
\begin{table}[!ht]
\resizebox{\linewidth}{!}{
    \centering
    \begin{tabular}{c|c|c|c|c}
    \midrule
        Datasets & $k$ &$lr$ & hidden layer $d$ & $\beta$ \\ \midrule
        100leaves &5& 0.01 & 64 & 0.00001 \\ 
        Animals &10& 0.001 & 128 & 0.00001 \\ 
        Flower17 &5& 0.0025 & 512 & 0.02 \\ 
        NGs &30& 0.0001 & 256 & 0.038 \\ 
        NoisyMNIST\_15000 &5 & 0.007 & 512 & 0.3 \\ 
        YaleB\_Extended & 5 & 0.04 & 256 & 0.08 \\ \midrule
    \end{tabular}}
    \caption{A brief description of the test datasets' parameters.}
    \label{ParaDescription}
\end{table}
\section{Parameter Sensitivity Analysis}
In Fig.~\ref{pa}, we present the comprehensive parameter sensitivity experiments conducted for our Multi-View Incremental Learning (MVIL) model, which underscore the robust nature of our model's performance. 
It can be seen that in datasets with a large number of classes, the smaller the value of $k$ is, the better the performance is; and most of the datasets show the phenomenon that the larger the number of hidden layer $d$ is, the better the performance is.

\section{Visualization of Node Representation}

Fig.~\ref{Tsne} shows the t-SNE visualization of node representations derived for NoisyMNIST\_15000.
The visualization underscores MVIL's proficiency in distinctly differentiating nodes, where a stark separation is observable between nodes exhibiting small intra-class distances and those featuring substantial inter-class distances, validating its effectiveness.

\end{document}